\documentclass{article}

\usepackage[dblblindworkshop, final]{neurips_2026}
\workshoptitle{Personalized, aligned, long-term memory for AI systems}

\usepackage{amsmath}
\usepackage{graphicx}
\usepackage[utf8]{inputenc} % allow utf-8 input
\usepackage[T1]{fontenc}    % use 8-bit T1 fonts
\usepackage{hyperref}       % hyperlinks
\usepackage{url}            % simple URL typesetting
\usepackage{booktabs}       % professional-quality tables
\usepackage{amsfonts}       % blackboard math symbols
\usepackage{nicefrac}       % compact symbols for 1/2, etc.
\usepackage{microtype}      % microtypography
\usepackage{xcolor}         % colors

\usepackage{enumitem}
\usepackage{xspace}

\title{Route to Remember: Adaptive Memory Selection for LLM Agents}

\author{
  Zeyu Fang\textsuperscript{1,2}\thanks{Corresponding Author: Zeyu Fang, joey.fang@gwu.edu},
  Jun Yin\textsuperscript{2},
  Yongsheng Mei\textsuperscript{2},
  Xu Han\textsuperscript{2},
  Chaitali Joshi\textsuperscript{2},
  Zhongkai Sun\textsuperscript{2},\AND
  Jie Hao\textsuperscript{2},
  Bin Guo\textsuperscript{2},
  Chengyuan Ma\textsuperscript{2},
  Benjamin Yao\textsuperscript{2},
  Chenlei Guo\textsuperscript{2}\\[0.5em]
  \textnormal{
    \textsuperscript{1}George Washington University
    \quad
    \textsuperscript{2}Amazon
  }
  }

\begin{document}

\maketitle

\begin{abstract}
The memory of production agents is increasingly becoming a bottleneck and is built
almost exclusively from text: key points are extracted from out-of-limit context and
retrieved into context at query time. Therefore, every new query incurs an additional cost that
grows with the accumulated history. Weight memory is a complementary alternative that internalizes
knowledge into model parameters and answers with no context at all, and has recently proven
effective in self-evolving agents for internalizing skills and behaviors. However, its application in episodic and
semantic memory, such as user-specific facts, remains comparatively unexplored, and a 
gap remains between earlier attempts at weight memory and text-based systems regarding recall performance.
We study how to make weight memory competitive as a long-term factual store by applying on-policy
self-distillation (OPSD). It trains a context-free student from a teacher, which is the same model
conditioned on the user's memory, together with a verify-then-regenerate router that lets weight and
text memory each cover the queries they handle best. On the HaluMem benchmark, OPSD matches or
surpasses strong text-memory baselines, including full-context prompting and retrieval systems, while reading
fewer tokens per query. We further conduct an extensive study that characterizes the
behavior of weight memory and analyzes how different training recipes affect its performance.
As agent harnesses move toward persistent, multi-session deployments, these
results position weight memory as a practical, low-cost component of agent memory and provide
insights into feasible training and evaluation protocols for combining it with retrieval.
\end{abstract}
% =====================================================================
\section{Introduction}

%Paragraph 1:
%we need a rigorous summary about the current trending. 
% 1. brief introduce the function and importance of memory in LLM agents/harness 
% 2. The current trend: All production agents like Codex and Claude are using text-based memory, and several text-based memory have been proposed (cite)
% 3. Limitation about the text-based memory 

% Paragraph 2:
% 1. Parametric memory, or weight-based memory is an alternative
% 2. Early study on that, yet not perform well when totally deprived of contexts 
% 3. Recent works about self-evolving agenttake better idea about this and is working great
% 4. Despite these, their works didn't focus on the memory system itself, especially for episodic/semantic facts; a large research blank here exists.

% Paragraph 3:
% the proposed methodology: Apply OPSD
% OPSD enhance the weight memory to outperform ..., yet it still face ...; So we think of a combination
% leads to the router design

% Paragraph 4:
% lots of experiments we did, and what we've found. Summary of the analysis and our conclusions

% Paragraph 5:
% itemize the contribution we have 

Persistent agents need to recall project conventions, user preferences, and earlier interactions across sessions.
Many memory systems store this information as text, then retrieve relevant entries into the prompt \citep{memgpt2023, hipporag2025, amem2025, memos2025}.
This design makes stored evidence inspectable but introduces retrieval and context-processing costs and depends on retrieving the relevant facts.

Storing memory in weights provides an alternative. If user history is internalized into
parameters, the following queries can be answered without retrieval or context, so the marginal cost of a query stops
depending on how long the dialogue is, and the cost is a training pass whenever memory is
written. Early attempts to use weights as memory, however, underperform once the model is deprived
of context at inference and are studied mostly for verbatim recall
\citep{loraasmemory2026, loraremembers2026}. Recent self-evolving agents revive the idea more
successfully, showing that on-policy distillation can internalize procedural skill into
weights \citep{opdevolver2026, kbsd2026}. But none of this work isolates the memory itself from the harness,
and the case that matters most for personalization is left open: the episodic and semantic memories built from timestamped sessions can arrive incrementally, update, or even sometimes contradict each other. The nearest evidence is a diagnostic of substrate asymmetry \citep{substrate2026}, which compares a per-user adapter against retrieval and concludes
that neither dominates and the two are complementary --- leaving unanswered how far a parametric
memory can actually be pushed on episodic and semantic recall and what it would take.

We address that question directly and apply on-policy self-distillation (OPSD): a per-user Low-Rank Adaptation (LoRA), referred to as the student, which is trained to reproduce, with no context at inference, the answers that the
same model provides with context in the prompt, referred to as the teacher. 
The distillation takes place on the student's own distribution, so the objective matches the deployment condition.
OPSD makes weight memory outperform every text baseline we test, but it requires extra training costs and inherits two
limits intrinsic to its parameters: it forgets older facts along with new updates, and it cannot
tell whether a fact was never recorded, so it confabulates instead of abstaining. Because weight and
text memory fail on different questions, we then ask whether they can combine. We adopt verify-then-regenerate (VtR): the parametric memory
answers first; those answers are audited against the text store, and only conflicted answers are
regenerated with the retrieved evidence, auditing the produced answer rather than judging from the query.

These claims rest on a broad experimental study on benchmarks for long-term memory such as HaluMem~\cite{chen2025halumem} and LoCoMo~\cite{maharana2024evaluating}, and four findings stand
out. First, OPSD reaches an overall QA accuracy of $59.9\%$, while ordinary Supervised Fine-Tuning (SFT) on identical data reaches $36.9\%$,
which is below every text baseline; the accuracy holds across model variant scales. Second, the three memory paradigms
differ less in accuracy than in how they degrade with question age: context truncates,
weights forget gradually, retrieval barely notices, and within weight memory, the learned habit of
declining survives while the content it protected decays. Third, the one deficit training does not
fix is inspecting absence, which can be traced to a single training-side
lever. The model's willingness to decline is attributable entirely to the ratio of refusal samples included in the training set. Fourth, VtR adds $6.2$ points over OPSD and improves five of six question categories,
whereas every other query-side routing signal we tried, including similarity, learned query features, model consistency, and answer confidence, can hardly predict whether the answer of weight-based memory is correct. Extensive analysis is conducted on these observations. 

We summarize our contributions as follows.
\begin{itemize}[leftmargin=*,itemsep=2pt,topsep=3pt]
  \item \textbf{An evaluation of OPSD for recent user-memory recall.}
    On the question set for eight HaluMem-Long users, OPSD reaches $59.9\%$, compared with $47.8\%$ for MemOS-style retrieval, $51.9\%$ for extracted context, and $51.3\%$ for raw dialogue.
    The advantage over retrieval is positive for all eight users, and the comparison with extracted context and retrieval holds at 4B, 9B, and 27B.
  \item \textbf{A characterization of retention and training effects.}
    Repeated questions distinguish context truncation, gradual forgetting in weights, and more stable retrieval.
    Our ablations identify training-objective and refusal-supervision effects within the tested recipes, while cost accounting separates query token usage from unmeasured deployment costs.
  \item \textbf{A routing study and VtR integration.}
    Query-side and confidence signals provide limited correctness information in our probes.
    Answer verification with text-based regeneration adds $6.2$ points over OPSD on native HaluMem questions, improving five of six categories while reducing memory-conflict accuracy.
\end{itemize}

% =====================================================================
\section{Related Work}
\label{sec:related}

\paragraph{Text memory systems.}
Text memory in deployed agents follows a typical loop: key points are extracted from the dialogue as natural-language statements, written to external storage, then retrieved by relevance when a new query arrives and injected into the prompt. 
Within that shared skeleton, systems differ mainly in how they organize and consolidate.
\citet{amem2025} maintain atomic notes with evolving cross-links, \citet{hipporag2025} build a knowledge graph, and \citet{thoughtretriever2026} store distilled reasoning rather than raw
text. A further line evolves the memory online from feedback \citep{liveevo2026, ace2026, atlas2026}. \citet{memos2025} generalize the picture into an
operating-system abstraction spanning plaintext, activation, and parameter memory.
However, these text-based systems inherit the same costs. Every query incurs retrieval latency and context tokens, and because the store grows while the window does not, retrieval quality
eventually becomes the binding constraint on accuracy. Surveys of the area
reach the same conclusion from the opposite direction, finding that consolidation, updating
, and forgetting remain far less developed than retrieval \citep{rethinkingmemory2025, memoryagentbench2026}, and that a longer
context window is not a substitute for a memory system \citep{agentmemsurvey2026}. We take \citet{memos2025} as our primary text-based system baseline.

\paragraph{Weight memory systems.}
The alternative writes new information directly into the model parameters. Editing methods localize a fact and overwrite the responsible feed-forward
weights \citep{rome2024}, which is precise but degrades after thousands of edits.
Others attach side feed-forward modules to a frozen backbone \citep{locas2026} or update fast
weights during inference \citep{inplacettt2026}, while
\citet{titans2025} learns a memory module updated by surprise-gated descent with an explicit
forgetting gate. Empirical analyzes of LoRA as a memory store \citep{loraasmemory2026, loraremembers2026} report that adapter capacity scales with rank while efficiency peaks at low rank, and that question-answer
formatted synthetic data substantially improves memorization, which is consistent with our own findings. Closest to our setting are methods using
distillation: \citet{sdpo2025} compress in-context feedback into parameters through KL
minimization, and the same idea appears in recent self-evolving agent work. \citet{opdevolver2026} distills a hindsight-conditioned teacher into a context-free student to internalize skills, and \citet{kbsd2026} uses self-distillation to calibrate search behavior. Our training method belongs to this
family, and what distinguishes it is the object rather than the optimizer: the teacher's privilege
is a recency-bounded window of the user's accumulated memory, the student is context-free at evaluation, and the target is a specific user's episodic and semantic memory rather than a skill or
a search policy.

\paragraph{Routing between memory systems.}
Routing is the natural response to the observation that not a single memory system can win all behavioral consistency, fact recall, and absence calibration at the same time\citep{substrate2026}. Existing routers on memories are mostly built within the same type, either text-based or parametric: \citet{wise2024} route a query between frozen and edited parametric memory by activation distance; \citet{loraasmemory2026} routes among multiple learned adapters through similarity; \citet{trustmargin2026} arbitrates the model's closed-book answer against a retrieved one by likelihood margin. We tested various signals for routing and finally designed a verify-then-regenerate mechanism for routing between two memory systems, making the memory reader closer in spirit to the write-time verification gate of \citet{atlas2026}.

% =====================================================================
\section{Method}
\label{sec:method}

To make weight memory competitive, we train it using OPSD~\citep{selfdistilled2026}: when a memory write is triggered, we convert the new dialogue contents into a question-answer training set and distill a
context-free student from a teacher that uses the same model but can see the context window. To combine the two memory systems
where each is strong, we then add a VtR reader that acts as a router: it
fact-checks the weight memory's answer against the text store, judges whether the store supports it,
and regenerates from retrieved context only when it does not.

\subsection{Data preparation}
\label{sec:data}

We call one assistant response a \emph{turn}, and a \emph{session} a sequence of turns spanning one
period of interaction. A user's history is then a sequence of sessions $S_1, \dots, S_T$, and the evaluation is conducted after each complete session through question answering, so memory is written once per session and read after every session.

Both memory systems we study consume the same intermediate representation for adding a session into memory: a base language model reads each session's dialog and extracts its key memory points, each written as one narrative sentence carrying its timestamp. A text memory system stores and retrieves those sentences directly; we instead convert them into training supervision, using the same base model for the augmentation
that produced the extraction. How that supervision is shaped matters more than how much of it there is. Continued pretraining on memory points or dialogue as raw text transfers poorly to question answering; what works is generating question-answer pairs in the same prompt template used at evaluation, and training with the loss masked to the answer. For each memory point we then generate three kinds of pair.
\emph{Paraphrases} ask for the same fact several ways, which is what later lets the student
generalize across phrasings rather than memorize one surface form. \emph{Counterfactuals} state the
fact wrongly and require a correction, which teaches the corrective form that questions carrying a
false premise demand. \emph{Unknowns} ask about attributes the session never recorded and require an
explicit refusal; these turn out to be the entire source of the model's willingness to decline, and
removing them is catastrophic in a way \S\ref{sec:char} quantifies.

\subsection{On-policy self-distillation of memory}
\label{sec:opsd}

We train a per-user LoRA adapter to answer without context, using the same base model
conditioned on the context window as the teacher. Denoting $\pi_\theta$ for the student and $\pi_{\mathrm{base}}$ for
the base model, the teacher is $\pi_{\mathrm{base}}(\cdot \mid q, M_t)$ and the student is
$\pi_\theta(\cdot \mid q)$ with no memory in the prompt. Here $M_t$ is not the entire accumulated
memory, which eventually exceeds any context budget, but a recency-bounded window over it:
the current session's memory points, always kept, plus as many of the most recent prior sessions as
fit a fixed budget, usually the context limit of the model. The student, being
a persistent adapter, keeps evolving across sessions, so it can retain
facts that no individual teacher context still contained, a point we return to when we ask why the
student can outperform its teacher (\S\ref{sec:main}). Training proceeds on-policy: for a question
$q$ drawn from the pool, the student samples its own answer $y \sim \pi_\theta(\cdot \mid q)$, the
teacher scores exactly those tokens, and we minimize the divergence between them on the sampled
trajectory,
\begin{equation}
\mathcal{L}(\theta) \;=\; \mathbb{E}_{q}\,\mathbb{E}_{y \sim \pi_\theta(\cdot \mid q)}
  \sum_{i} \mathrm{KL}\!\left(\pi_{\mathrm{base}}(\cdot \mid q, M_t, y_{<i})
  \,\middle\|\, \pi_\theta(\cdot \mid q, y_{<i})\right).
\end{equation}
Sampling from the student rather than the teacher is what keeps the objective on-policy, and it
matters for both a mundane reason and a theoretical one: an off-policy variant, like supervised learning over the teacher's answers, suffers from exposure bias and degenerates into repetition loops, scoring well
below the on-policy version.

Two design choices proved decisive for training robustness. The first is
that the teacher should not be replaced by the student: Using the accumulating student snapshot from the
previous session as a teacher, the natural self-distillation choice lets errors compound across
sessions, and the forgetting curve collapses late in the horizon. The second
is that the question pool should stay fixed or grow only within a reasonable range. Expanding it
with an adversarially trained questioner helps in a single-session probe but hurts across many
sessions, where the generated questions can drift onto unrelated topics and dilute the native facts
the model needs to retain. Since memory is written once per session and read many times, we keep the
adapter persistent across sessions, so each session's training accumulates on top of all previous
ones. Adding a regularization term may be a solution for both problems, however it can introduce extra costs to tune the hyperparameter. 

\subsection{Verify-then-regenerate as the router}
\label{sec:vtr-method}

Weight memory answers fluently whether or not it knows, and its one systematic blind spot, absence,
is exactly what a text store makes visible. That asymmetry suggests auditing the weight memory's
answer rather than choosing between the two systems in advance based on the query, and \S\ref{sec:route} shows why the
alternative fails. Given the student's answer $\hat{y}$ to question $q$ at session $t$, we retrieve
from the text store twice, once with $q$ as the query and once with $\hat{y}$, taking the top $k$ of
each. A verifier then judges $\hat{y}$ against the answer-anchored evidence as supported,
conflicting, or unsupported. Because it only has to emit one of three labels, the verifier is a
cheap classifier rather than a second LLM agent. If the verdict is supported, the weight memory's
answer stands. Otherwise, the base model, with no adapter, re-answers from the union of both retrieved evidence sets, each memory point annotated with a timestamp, under an instruction to reply properly with the retrieved evidence.

Three details carry the result: (1) Where the evidence comes from matters. We retrieve from the
store managed by the text memory system, not from the weight memory's raw training data. This is
what allows the text-based memory system to handle updates, which a case weight memory handles poorly on its own, because a later session's training set can overwrite an earlier fact, for example, a user's changing physical address. More specifically, in this experiment, we retrieve against a \emph{merged} store, in which superseded values have been integrated into the previous related statements or totally replaced them; without merging, a
third of all verdicts come back as conflicts, while merging reduces that to about one percent.
(2) Retrieving with a generated answer is not the same operation as retrieving with
the question because it inserts a step of semantic reasoning. Weight memory produces a plausible
answer, which maps the question into a region of the space where memory points themselves live;
matching against that region is a better-posed retrieval problem than matching a question against
statements that are not phrased as questions, and it cuts retrieval noise accordingly. The two keys
are also deliberately kept distinct since verification and regeneration need different evidence:
the audit should see what was actually claimed, while the regenerator should see what the question
makes answerable. The two sets overlap only partially, and the answer-anchored half contributes
evidence that the question alone does not surface.
(3) Only the base model is used for regeneration; therefore, it is adapter-free by design. The fallback path is purely text-based and thus cannot inherit the confabulations that the audit aims to catch.

% =====================================================================
\section{Experiments}
\label{sec:exp}

\subsection{Setup}
\label{sec:setup}

\paragraph{Benchmarks.}
HaluMem \cite{chen2025halumem} is our primary benchmark. It has in total dialogue data of $20$ distinct users, each with $109$--$136$
sessions and roughly a million tokens per use. We choose it because it evaluates
after every session rather than once after the whole dialogue, and because its question taxonomy separates the phenomena we care
about: basic fact recall, memory conflict, dynamic update, multi-hop inference, generalization and
application, and memory boundary.  That last category is where the question is asking for out-of-distribution facts, and the agent should refuse to answer. We additionally tested our method over LoCoMo \cite{maharana2024evaluating}, a
multi-session dialogue benchmark, to check that the result is not an artifact of HaluMem's
generation process; it has no native notion of question age, so the retention protocol remains
HaluMem-only.

HaluMem originally asks each question only in the session that first makes it answerable, and that native
distribution is what all our headline numbers report. To study retention, we add a
supplementary protocol that re-asks every question $20, 40, \dots, 80$ sessions later, which we report the results per each age to trace each memory system's forgetting curve
(\S\ref{sec:char}). 
We do not average the two, because the re-asks outnumber the native questions
roughly two to one and the memory systems decay at different rates, so a pooled mean reflects the
mixing ratio as much as the systems.

\paragraph{Baselines.}
Three text memory systems span the cost spectrum: MemOS-style retrieval over a database with top-$k$ fetch, the full extracted memory in context, and the raw dialogue truncated to the context
limit. We use MemOS because it is the strongest text memory system in HaluMem's own evaluation.
Ground-truth memory points in context, provided by the benchmark, serve as an oracle and is denoted as GT points. On the weight
side we compare against SFT on identical augmented data in the style of
\citet{loraasmemory2026}, which isolates the contribution of the distillation objective.

\paragraph{Protocol.}
One language model performs every text operation in the pipeline, including memory-point
extraction, text-store management, training-data preparation, and question answering, so
differences between systems are attributable to the memory rather than to the reader. We use
Qwen3.5-9B for the main comparison and repeat it at a smaller scale to check that the effect is not
scale-specific (Appendix~\ref{sec:app-scale}). Claude Sonnet 4.6 serves as the judge, labelling each
response correct, hallucinated, or omitted, with no partial credit; we report Correct\% throughout,
and the absence of partial credit is worth keeping in mind for questions whose ground-truth answer is a
terse label. We report paired McNemar tests over matched
questions, $95\%$ confidence intervals from a bootstrap clustered on users, and per-user
differences, so that no claim rests on a single user's behaviour.
Full training and retrieval hyperparameters are given in Appendix~\ref{sec:app-impl}.

\subsection{OPSD-based weight memory outperforms text memory}
\label{sec:main}

Table~\ref{tab:cat} gives per-category accuracy for every method on HaluMem's native question set
($n{=}1438$); Table~\ref{tab:cost} gives the cost of each. We read the two together: weight memory
is at once the most accurate non-oracle system and the cheapest to query, at a higher write cost.

\begin{table}[t]
\centering
\caption{Per-category accuracy (Correct\%) on HaluMem's native questions. GT points is an oracle
(upper bound if memory point extraction is ideal); VtR is the verify-then-regenerate reader (\S\ref{sec:route}). Overall gaps
vs.\ OPSD are significant (paired McNemar $p{<}10^{-9}$).}
\label{tab:cat}
\small
\setlength{\tabcolsep}{4pt}
\begin{tabular}{llccccccc}
\toprule
& system & Basic & Conflict & Dynamic & Multi-hop & General. & Boundary & \textbf{Overall} \\
& & \footnotesize(311) & \footnotesize(304) & \footnotesize(94) & \footnotesize(79) & \footnotesize(311) & \footnotesize(339) & \footnotesize(1438) \\
\midrule
oracle & GT points          & 74.3 & 77.3 & 57.4 & 40.5 & 34.4 & 77.0 & 64.0 \\
\midrule
text   & Extracted (dump)   & 50.2 & 69.7 & 18.1 & 16.5 & 35.4 & 70.5 & 51.9 \\
text   & Raw dialogue       & 50.5 & 71.1 & 18.1 & 17.7 & 31.5 & 69.3 & 51.3 \\
text   & MemOS (retrieval)  & 33.8 & 58.6 & 26.6 & 10.1 & 22.5 & 89.1 & 47.8 \\
\midrule
weight & SFT                & 9.0  & 53.3 & 3.2  & 7.6  & 10.0 & 88.8 & 36.9 \\
weight & \textbf{OPSD}      & 54.3 & \textbf{76.0} & 29.8 & 24.1 & 44.1 & 81.7 & 59.9 \\
\midrule
both   & \textbf{VtR}       & \textbf{59.2} & 66.1 & \textbf{35.1} & \textbf{36.7} & \textbf{53.4} & \textbf{99.4} & \textbf{66.1} \\
\bottomrule
\end{tabular}
\end{table}

\begin{table}[t]
\centering
\caption{Usage per method. \emph{Write} is the one-time cost of building a user's memory (LLM
in/out tokens); \emph{read} is the per-query evaluation cost (tokens and GPU-h). Storage and weight
methods share the extraction pass ($30.2$M/$12.0$M) plus a method-specific term (merge for MemOS,
QA augmentation for the weight methods). All experiments are done on NVIDIA H200 GPU.}
\label{tab:cost}
\small
\begin{tabular}{lrrrrr}
\toprule
& \multicolumn{2}{c}{write tokens} & \multicolumn{2}{c}{read tokens} & read \\
\cmidrule(lr){2-3}\cmidrule(lr){4-5}
method & in & out & in & out & GPU-h \\
\midrule
Raw Dialogue      & ---            & ---           & $1.07$B & $0.46$M & 23.4 \\
Extracted (dump)  & $30.2$M        & $12.0$M       & $1.03$B & $0.39$M & 22.5 \\
MemOS (retrieval) & $30.2{+}19.6$M & $12.0{+}1.3$M  & $1.7$M  & $0.05$M & 0.9 \\
\textbf{OPSD}     & $30.2{+}76.5$M & $12.0{+}31.4$M & $326$K  & $0.06$M & 1.0 \\
\textbf{OPSD + VtR} &  $30.2{+}96.1$M & $12.0{+}32.7$M & $1.3$M & $0.09$M & 1.8 \\
\bottomrule
\end{tabular}
\end{table}

\paragraph{Weight memory is more accurate, and the recipe is what makes it so.}
OPSD reaches $59.9\%$, above every text baseline, including the two full-context systems that
hold the entire history in the prompt ($51.3\%$ and $51.9\%$) and the MemOS retriever ($47.8\%$).
The margin over retrieval is $12.1$ points, with a user-clustered confidence interval of
$[9.4, 14.8]$, positive for all eight users individually, from $7.1$ to $18.1$ points; on matched
questions, OPSD is uniquely correct on $309$ questions against retrieval's $136$. We take the comparison
against the full-context base model to be the more informative one: even though the context is not shown, the OPSD method still does better relying purely on its weights.
The advantage is broadly distributed --- OPSD leads on five of six question categories --- and it
holds across model size, leading every baseline at 4B and 27B (Appendix~\ref{sec:app-scale}). It is,
nonetheless, an accuracy that is concentrated on recent material and blind to what was never
recorded: weight memory forgets as facts age and confabulates when asked for an attribute the user
never gave, the two failure modes we characterize in \S\ref{sec:char}. The latter is exactly the memory boundary, the one category OPSD loses, where the answer is an abstention, and a text system can simply observe
that its retrieved list contains nothing relevant (\S\ref{sec:absence}); this is also the only
category propping up the otherwise weak text baselines, and removing it widens OPSD's overall margin
over retrieval from $12.1$ to $18.0$ points.

That this accuracy is a property of the training recipe rather than of parametric storage is
shown by the weakest row in Table~\ref{tab:cat}: supervised fine-tuning on the identical augmented
data reaches only $36.9\%$, below every text baseline, even though its stored representation and
training data match OPSD's and only the objective differs. Early results discouraging about
weight memory were obtained almost entirely with supervised fine-tuning
\citep{substrate2026, loraasmemory2026}; our comparison indicates that on-policy self-distillation,
not SFT, is the appropriate objective for internalizing episodic/semantic memory, and that the
medium was never the constraint those results attributed it to. Appendix~\ref{sec:app-recipe}
traces how the OPSD configuration was reached.

\paragraph{Weight memory is costly to write but cheap to query.}
Table~\ref{tab:cost} shows that the two families invert where their cost falls. A text system
writes for free but reads an entire context on every query, spending $1.7$M to over a billion input
tokens across the evaluation; weight memory front-loads the cost into writing --- extraction, QA
augmentation, and a distillation pass not billed in tokens --- and then reads only the query tokens,
four orders of magnitude fewer than full context and a factor of five below retrieval. Because
writing is paid once per stored memory while reading recurs, weight memory pays off precisely when
the same memory is queried many times: its natural use case is internalizing a common fact or skill
that thousands of users will each query, rather than a single user's transient history. The
crossover is quantifiable from Table~\ref{tab:cost}. Against full context, OPSD's roughly $150$M
extra write tokens are repaid after about $600$ queries per stored memory, since each query saves
about $2.4\times10^{5}$ input tokens. Against retrieval the break-even is far higher, on the order
of $10^{5}$ queries, because retrieval is already cheap to read (about $310$ input tokens saved per
query against a $57$M-token extra write): weight memory dominates full context almost immediately
but only overtakes retrieval on cost under very heavy, highly shared query load. VtR keeps weight
memory's read cost on the $38\%$ of queries the gate accepts and pays a retrieval-plus-regeneration
read on the rest, so its per-query cost sits between OPSD and MemOS (Table~\ref{tab:cost}).

\paragraph{Why the student can beat its teacher.}
OPSD ($59.9\%$) also exceeds the teacher configuration it distils from --- the base model reading
extracted memory in context ($51.9\%$). We offer three possible explanations and leave further analysis for future work:
(i)~\emph{Consolidation:} training over many paraphrases of a fact averages over the extraction
noise that any single retrieved statement carries, in the psychological sense of consolidation.
(ii)~\emph{No in-context search:} the student never has to locate its answer among hundreds of
distractor statements, which is where the full-context memory systems visibly fail, their
hallucination rate sitting at $33$--$34\%$.
(iii)~\emph{Persistence past the teacher's window:} because the teacher sees only a recency-bounded
context while the adapter persists, the student can retain facts that have since dropped out of any
single teacher window rather than re-finding them at query time. 

\subsection{Characterizing weight memory as long-term memory}
\label{sec:char}

\paragraph{Retention: content decays, habits do not.}
\begin{figure}[t]
\centering
\begin{minipage}[t]{0.45\linewidth}
\centering
\includegraphics[width=\linewidth]{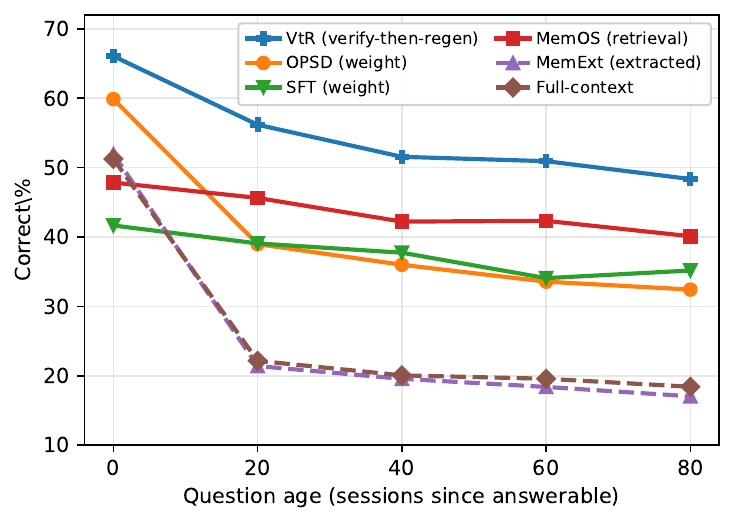}
\caption{HaluMem-Long retention---Correct\% by question age; context truncates, weight decays,
retrieval stays flat. }
\label{fig:age}
\end{minipage}\hfill
\begin{minipage}[t]{0.42\linewidth}
\centering
\includegraphics[width=\linewidth]{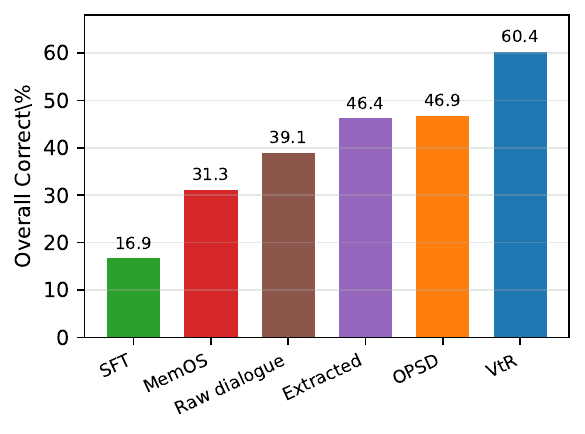}
\caption{LoCoMo overall accuracy---the ranking of Fig.~\ref{fig:age} reproduces on an independent
benchmark. Per-category in Table~\ref{tab:app-locomo}.}
\label{fig:locomo}
\end{minipage}
\end{figure}

Figure~\ref{fig:age} separates the three paradigms by failure modes. Context
methods fall off a cliff when the relevant session leaves the window, and the more compressed the
representation the later the cliff arrives: raw dialogue collapses by age $40$, extracted memory
holds until $60$. Weight memory decays smoothly instead, from $59.9$ to $32.4$ by age~$80$.
Retrieval is nearly flat, from $47.8$ to $40.1$, because a top-$5$ fetch is indifferent to how old a
fact is. This is the precise scope condition on \S\ref{sec:main}: weight memory's advantage is
concentrated on recent material, and retrieval overtakes it beyond roughly twenty sessions. It also suggests the write
policy a deployed system should adopt, distilling recent sessions into weights while leaving the
tail to retrieval.

Within weight memory, forgetting is strikingly non-uniform. Between age $0$ and age $100$ the
memory-boundary category is flat, drifting from $81.7$ to $88.1$, while every content category
collapses. The
model does not forget that it should be careful; it forgets what it knew, and keeps answering
anyway.

\paragraph{Second benchmark: LoCoMo.}
To check that the ranking is not an artifact of HaluMem's generation process, we port LoCoMo, a
multi-session dialogue benchmark independently constructed. LoCoMo asks all questions once after the
full conversation, so there is no notion of question age; we therefore pool every session into a
single memory and report end-of-conversation accuracy per LoCoMo category (single-hop, temporal,
multi-hop, open-domain, and adversarial, the last being unanswerable-from-context questions that
should be refused). Figure~\ref{fig:locomo} (right) shows overall accuracy; the full per-category
breakdown is in Table~\ref{tab:app-locomo} (appendix).

The HaluMem ranking reproduces. OPSD matches extracted text memory on the content categories
(single-hop $70.2$ vs $69.3$, temporal $61.3$ vs $61.7$, multi-hop $33.5$ vs $33.3$) and edges it
overall ($46.9$ vs $46.4$), while both crush SFT ($16.9$); raw full-context again loses on temporal
questions ($7.2$), where a whole conversation in the window obscures when each fact was held.
The one category where every paradigm fails is adversarial: pure weight and pure text memory both
sit near the floor ($2.5$--$10.6$), because neither reliably refuses an unanswerable question. This
is the same absence-inspection failure as HaluMem's memory-boundary category
(\S\ref{sec:absence}), and it is exactly what the router repairs: verify-then-regenerate lifts
adversarial accuracy to $73.1$ and, without hurting the categories weight memory already answers,
takes the overall best at $60.4$, the same $+13$--$15$pt hybrid gain we see on HaluMem, now on an
independent benchmark.

\paragraph{Absence is not inspectable in weights.}
\label{sec:absence}
The one category where text memory beats weight memory is the one where the correct
answer is that nothing was recorded. The mechanism is not knowledge but inspectability: a text
system reads a retrieved list, finds no such field, and reports its absence, whereas parameters
offer nothing to inspect and instead supply a plausible value. The inventions are specific enough
to be diagnostic, with concrete salaries and body weights produced for attributes the dataset never
contained, inconsistently across ages for the same question. This is precisely the confabulation
\citet{substrate2026} identify for single-turn personalization, reproduced here in a multi-session
setting. About a fifth of the boundary ``hallucination'' verdicts are in fact valid abstentions
scored wrong while a near-identical text refusal was scored correct, so the true fabrication rate is
roughly $80\%$ of the stated one; this is one instance of the rubric artifacts catalogued in
Appendix~\ref{sec:app-weak}.

\paragraph{What the training recipe contributes.}
The recipe rather than the medium is what makes weight memory work, and we ablate it along every
axis we could vary; the full sweeps are in Appendix~\ref{sec:app-results} and we summarize it here. (1) \emph{Objective} is the dominant lever: on-policy self-distillation ($59.9\%$) beats both
supervised fine-tuning on identical data (Table~\ref{tab:cat}, $36.9\%$) and an off-policy variant that
teacher-forces the teacher's raw text rather than the student's own samples, and full-parameter
fine-tuning does not improve on the LoRA adapter (Appendix~\ref{sec:app-recipe}). (2) \emph{Teacher
construction} is next: decoupling the teacher from the accumulating student to a frozen base model on
capped cumulative memory removes a late-session collapse and is worth roughly $+14$ points over the
horizon, while curriculum expansion of the training-QA pool, helpful in a single session, reverses
online and is dropped (Appendix~\ref{sec:app-recipe}). (3) \emph{Training data} matters more than
capacity: unknown-answer augmentation is solely responsible for the abstention ability. Removing it
collapses the memory boundary from $88\%$ to $1\%$ and costs $14$--$16$ points overall, with hallucination
rising to $60\%$—while doubling the unknown/counterfactual budget or adding composite-QA augmentation
is at best neutral (Appendix~\ref{sec:app-aug}). (4) \emph{Anti-forgetting} replay of cached answers helps
the tail but craters dynamic update, because re-supervising a stale answer re-teaches a superseded
value (Appendix~\ref{sec:app-recipe}). (5) By contrast  \emph{capacity} is nearly inert: adapter rank is
flat from $16$ to $128$ (best at $64$, $128$ mildly overfitting) and training length is non-monotonic
(Appendix~\ref{sec:app-rank}). Across all of these it is the objective and the training-QA set, not the
size of the store, that determine what weight memory can do.

\subsection{Route design}
\label{sec:route}

Since the two memory systems fail on different questions, disagreeing on $445$ of $1438$, an oracle
choosing the better one per question would reach $69.3\%$ against OPSD's $59.9\%$. The
natural way to claim that prize is to inspect the query and decide which store to trust, and we
report four attempts, all negative, because together they explain the design of
\S\ref{sec:vtr-method}.

Query-memory similarity carries no signal at all: correct and incorrect answers have
indistinguishable top-$1$ similarity distributions ($0.735$ against $0.741$ for weight memory,
$0.737$ against $0.737$ for retrieval), so every threshold degenerates into always choosing the
better system. A classifier over query embeddings, similarity features and cue tokens gains at most
$0.9$ points, and raw query embeddings actively hurt: whether the weights happen to have
memorized a fact is not legible in the question. Confidence in the generated answer looks
promising until it is deconfounded: per-token confidence correlates at $0.65$ with answer length,
and length correlates with correctness with opposite signs across categories, positively where the
right answer is a verbose refusal and negatively where a rambling answer signals a miss, so a
single global probe hides two opposing effects. Controlling for that with a length-constant probe, which
teacher-forces a fixed refusal string of identical token length for every question, gives a
genuinely clean detector, with an area under the curve of $0.918$ for answerability and negligible
correlation with answer length. But it detects answerability rather than correctness: against the
event that the model is wrong, the same probe is at chance ($0.516$). Routing on it yields a single
point, and under the retention protocol it collapses entirely, sending essentially every question
to retrieval once the threshold drifts.

\begin{figure}[t]
\centering
\includegraphics[width=0.86\linewidth]{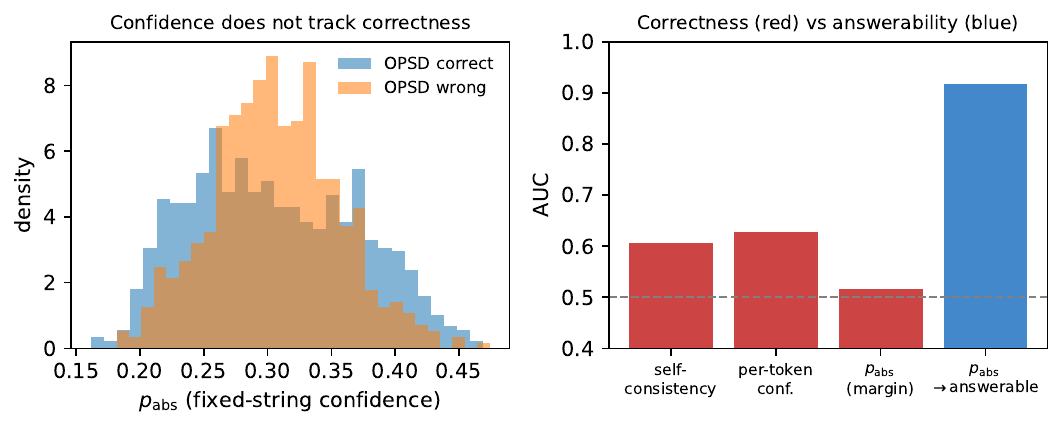}
\caption{\textbf{Left:} $p_\mathrm{abs}$ has near-identical distributions for correct vs.\ wrong
age-$0$ OPSD answers (means $0.305$/$0.306$), so no threshold separates them. \textbf{Right:}
query-side signals predict \emph{correctness} at chance (red, AUC${\approx}0.5$) but $p_\mathrm{abs}$
predicts \emph{answerability} at $0.918$ (blue): confidence knows if a question is answerable, not if
the answer is right. Details in Appendix~\ref{sec:app-conf}.}
\label{fig:router-signals}
\end{figure}

The common thread (Figure~\ref{fig:router-signals}) is that all four signals are functions of the
question and the model's internal state, and none of them observes the answer that was actually
produced. Verification does, which is
why VtR takes the form it does.

VtR reaches $66.1\%$, an improvement of $6.2$ points over OPSD
($p = 2 \times 10^{-6}$, user-clustered interval $[3.6, 8.6]$) and $18.2$ over the best text
baseline, and unlike any of the selection mechanisms above it improves at every question age
(Figure~\ref{fig:age}). The gate keeps $37.9\%$ of answers and regenerates the rest. Its behaviour
is category-dependent in a way that matches the per-category picture (Table~\ref{tab:regen}, Appendix~\ref{sec:app-vtr}): it
improves five of six categories, and every memory-boundary question is flagged, correctly, since
nothing in the store supports an unanswerable question, and the regenerated answer is a refusal,
which lifts that category to $99.4\%$.

% One category regresses, and its diagnosis is instructive. Memory-conflict questions embed a false
% premise and expect a correction rather than an answer (``no, he began a sabbatical, not a
% retirement''), and this is the category where weight memory has its largest lead over retrieval,
% because a fact consolidated into parameters resists a false premise stated in the prompt. The
% regeneration prompt's abstention instruction pushes the base model toward declining instead of
% asserting the correction.
% NOTE (not for print): suppressing regeneration on this category recovers the loss, but that value
% equals the per-category keep/regenerate oracle, so it cannot be reported as a pipeline score.
% The principled fix is a false-premise clause in the regeneration prompt; run before camera-ready.

% =====================================================================
\section{Conclusion}

Today's agents keep a user's memory in text and pay for it on every turn. We showed that this memory can
instead live in a model's weights by training with OPSD. Parametric memory should be held back not by the model or data but by the training protocol, since SFT on the same data has distinct performances. The characteristic study conducted on weighted-based memory demonstrates that it excels with recent material but forgets as facts age and cannot tell when a fact was never recorded; the regimes where a text store still wins. Our verify-then-regenerate
router lets the weight memory answer and then checks that
answer against the text store. For an agent
, this suggests a concrete design: consolidate a user's recurring or widely shared knowledge
into weights while keeping a text store for the recent and the absent, and
compose the two by verifying answers. 

The broader opportunity is that weight memory changes what an agent can afford to remember. Because
the write cost is paid once and the read cost is negligible, the natural unit to internalize is not a
single user's history but a pattern mined across many of them: a workflow that thousands of users
repeat, a tool-calling convention, a recurring preference, or an interaction style. Distilling such
shared regularities, with per-user memory layered on top, would let a harness carry common competence in its weights and reserve retrieval for what is genuinely individual and
current. This also points beyond text. The same objective should apply to procedural and skill
memory, to tool use, and, with a multimodal teacher, to knowledge that never had a clean textual
form, so that an agent accumulates capability rather than merely facts, which marks the future direction of this research.

\bibliography{related_works}
\bibliographystyle{unsrtnat}

\newpage

\appendix
% =====================================================================
% Supplementary results appendix. Self-contained: no preamble, no
% \appendix, no \begin{document}. \input after \appendix in the main file.
% Uses booktabs; every label prefixed tab:app-* / fig:app-* to avoid
% collisions with the main-text labels.
% =====================================================================

\section{Additional results}
\label{sec:app-results}

% This appendix collects the finer-grained results that the main text summarizes or references
% under the page limit.
% Where a supplementary table disagrees with the main text, the main text uses the age-$0$ native
% slice and this appendix reports the pooled all-age slice; the two are never averaged
% (\S\ref{sec:setup}).

\subsection{Implementation details}
\label{sec:app-impl}

All conditions use Qwen3.5-9B as the reader; the memory-point extraction, text-store management,
training-data augmentation, and question answering are performed by this same model. OPSD trains a
per-user LoRA adapter of rank $64$ ($\alpha{=}128$) for $30$ rounds per session, sampling $60$
on-policy questions per round; the adapter is persistent across sessions. The teacher is the same
base model conditioned on a recency-bounded window of the user's memory, capped at $12{,}000$
characters (current session always kept, whole older sessions dropped to fit). Retrieval, for both
the MemOS baseline and the verify-then-regenerate router, uses Qwen3-Embedding-0.6B with cosine
top-$5$ fetch. The judge is Claude Sonnet 4.6 with a correct/hallucinated/omitted rubric and no
partial credit. These settings are fixed across the experiments below unless a subsection states
otherwise.

\subsection{Model scale: the effect is not specific to Qwen3.5-9B}
\label{sec:app-scale}

The main comparison uses Qwen3.5-9B; we repeat it at 4B and 27B so the ranking is not read as a
single-model artifact. Table~\ref{tab:app-scale} collects results over all three scales. OPSD leads every text
baseline at both sizes, and supervised fine-tuning on the identical data is the weakest condition by a wide margin at 4B and 9B but rises steeply
with model size while distilled OPSD stays high, so the gap between the plain and the distilled recipe closes as the base model strengthens. Besides, the advantage of OPSD over the text-based model strengths from $1.2\%$ at 27B to $8.0\%$ at 9B and $9.4\%$ at 4B, indicating that the performance of smaller based model can be enhanced more significantly with OPSD.

\begin{table}[b]
\centering
\caption{Model-scale comparison over full question
set used in the main text; OPSD leads every text baseline at
every scale on the full-horizon set; SFT rising steeply with model size while distilled OPSD stays high, so the recipe gap closes as the base model strengthens. Besides, the advantage of OPSD over its teacher (Extracted) grows along with smaller models.}
\label{tab:app-scale}
\small
\begin{tabular}{lccc}
\toprule
method/model scale & 4B & 9B & 27B \\
\midrule
OPSD (weight)      & \textbf{57.4} & \textbf{59.9} & \textbf{62.6} \\
Extracted (dump)   & 48.0 & 51.9 & 61.4 \\
MemOS (retrieval)  & 45.6 & 47.8 & 47.4 \\
SFT (weight)       & 15.9 & 36.9 & 58.3 \\
\bottomrule
\end{tabular}
\end{table}

\subsection{LoRA rank sweep}
\label{sec:app-rank}

Table~\ref{tab:app-rank} sweeps adapter rank for the weight memory on the pooled all-age set.
Accuracy is nearly flat from $r{=}16$ to $r{=}64$, peaks at $r{=}64$, and $r{=}128$ regresses with
the highest hallucination rate, consistent with mild overfitting rather than a capacity ceiling.
Rank is therefore not the operative variable in this range, which is why the main text fixes
$r{=}64$ and treats the training objective, not adapter size, as the lever
(\S\ref{sec:main}, \S\ref{sec:char}).

\begin{table}[t]
\centering
\caption{LoRA rank sweep (HaluMem-Long, pooled $n{=}4428$), Correct/Halluc/Omit\%. Alpha $=2r$,
$100$ steps.}
\label{tab:app-rank}
\small
\begin{tabular}{lccc}
\toprule
rank & Correct\% & Halluc\% & Omit\% \\
\midrule
$r{=}16$  & 37.9 & 31.2 & 30.8 \\
$r{=}32$  & 39.2 & 28.7 & 32.1 \\
$r{=}64$  & \textbf{39.5} & 29.9 & 30.6 \\
$r{=}128$ & 37.2 & 34.7 & 28.1 \\
\bottomrule
\end{tabular}
\end{table}

\subsection{Augmentation ablations}
\label{sec:app-aug}

The training data, not the storage medium, determines what weight memory can do, and the augmentation
ablations locate the effect precisely. Table~\ref{tab:app-aug} reports the decisive one: removing
the synthetic unknown-answer pairs (the \texttt{u0} column) collapses Memory Boundary from roughly
$88\%$ to under $4\%$ and costs $14$--$16$ points overall, with hallucination rising to about
$60\%$. The model's entire ability to abstain is learned from those refusals; it is not a free
byproduct of parametric storage. This is the quantitative backing for the training-recipe paragraph
of \S\ref{sec:char}.

\begin{table}[t]
\centering
\caption{Unknown-augmentation ablation (HaluMem-Long, pooled $n{=}4428$), Correct\%.
\texttt{n5u1c1} keeps the unknown pairs, \texttt{n5u0c1} removes them. The entire loss is
concentrated in Memory Boundary.}
\label{tab:app-aug}
\small
\setlength{\tabcolsep}{4pt}
\begin{tabular}{lcccccccc}
\toprule
condition & Basic & Conflict & Dynamic & Multi-hop & General. & Boundary & \textbf{Overall} & Halluc\% \\
\midrule
$r64$, \texttt{n5u1c1} (base) & 14.6 & 56.8 & 7.1 & 10.6 & 11.8 & \textbf{88.4} & \textbf{39.5} & 29.9 \\
$r64$, \texttt{n5u0c1}, s100  & 24.6 & 55.9 & 21.4 & 19.2 & 17.4 & 0.8  & 23.6 & $\sim$60 \\
$r64$, \texttt{n5u0c1}, s200  & 28.8 & 56.6 & 22.7 & 18.4 & 20.6 & 3.4  & 25.9 & $\sim$60 \\
\bottomrule
\end{tabular}
\end{table}

Two further augmentation findings are directional (small categories) but
consistent. First, doubling the unknown and counterfactual budget (\texttt{u2c2} vs \texttt{u1c1})
is net-negative: on the HaluMem-Medium session probe it trails \texttt{u1c1} by $3$--$4$ points at
every training length, buying about $+5$ points on Memory Boundary while collapsing Generalization
($27\rightarrow18$) and Multi-hop ($33\rightarrow22$). Second, an embedding-graph composite-QA
augmentation ($u1c1cmp1$) reaches $50.0\pm1.9\%$ at $200$ steps on the same probe, its best cell,
and accuracy declines monotonically with more steps for both augmentation levels, indicating
overfitting past roughly $200$ training steps rather than an undertraining regime. These two results come
from a smaller HaluMem-Medium probe ($8{\times}10$ users, $196$ questions, training with SFT) and are reported as
directional; they motivate the "capacity is not the operative variable" claim of \S\ref{sec:char}
but should not be read as precise.

\subsection{Training objective and anti-forgetting}
\label{sec:app-recipe}

The following recipe results are directional---they explain how the OPSD configuration in
the main text was arrived at. \emph{Objective.} On the session-$0$ probe the objective ladder is
monotone: an off-policy variant that teacher-forces the teacher's raw answer text reaches $47\%$
and suffers repetition/degeneration from exposure bias, whereas the on-policy formulation (the
student samples its own answer and the teacher scores those tokens) reaches $61\%$ with clean
generations; supervised fine-tuning on the same data is the weakest and is the row reported in the
main comparison (Table~\ref{tab:cat}). Full-parameter fine-tuning of the base model, on the
value-only pipeline, does not beat the rank-$64$ LoRA adapter and is far more expensive, so we keep
the adapter. \emph{Teacher construction.} The dominant fix over the multi-session horizon is
\emph{teacher decoupling}: replacing the accumulating-student teacher (which itself degrades over
sessions) with a frozen base model reading capped cumulative memory lifts the full run from about
$51\%$ to about $65\%$ on the first-$10$-session probe ($+14$ points), with the largest gains on
Generalization ($20\rightarrow50$) and Basic Fact ($56\rightarrow78$), and removes the late-session
collapse. Curriculum expansion of the training-QA pool, which helped on the single-session probe
(reaching the $\sim$80\% teacher ceiling), \emph{reverses} online: growing the pool to $1000+$
generated questions dilutes the native facts, so the best multi-session configuration is base
teacher with a fixed cache pool and no expansion. A more aggressive questioner is worse still:
a learned questioner trained with a GRPO objective to maximize the student--teacher KL divergence,
that is, to ask whatever the student is currently worst at, reaches only $26\%$ on the session-$0$
probe against on-policy OPSD's $61\%$, because the difficulty-chasing reward collapses the generated
pool onto a single question type and starves it of diversity. The lesson across both variants is
that the questioner should broaden coverage of the recorded facts, not maximize hardness. On the full long horizon, an SFT-replay
anti-forgetting term (rehearsing cached answers to old questions) helps the far tail
(age-$100$ $27\rightarrow32\%$) but craters Dynamic Update, because re-supervising a cached answer
re-teaches a value that has since been superseded, a concrete instance of weight memory and
volatile facts interacting badly. Round count is non-monotonic with $30$ rounds the sweet spot.
This is the mechanism behind the retention-versus-freshness tension in \S\ref{sec:char}.

\subsection{Full retention detail}
\label{sec:app-retention}

Breaking OPSD's retention curve (Figure~\ref{fig:age} in the main text) into the six question
categories makes the non-uniform-forgetting claim of \S\ref{sec:char} concrete: Memory Boundary is
flat with age (the abstention habit persists, drifting $81.7\rightarrow88.1$) while every content
category collapses toward the floor, Generalization and Multi-hop reaching $0.0$ by age $100$. The
abstention behavior is a habit that survives; the knowledge is what decays. Table~\ref{tab:app-cho}
gives the accompanying Correct/Halluc/Omit trajectory: as OPSD forgets, hallucination climbs from
$33.0$ to over $50\%$ while omission barely moves, so the model becomes confidently wrong rather than appropriately cautious.

\begin{table}[t]
\centering
\caption{OPSD verdict mix by question age (auxiliary protocol). Hallucination, not omission, grows
with forgetting.}
\label{tab:app-cho}
\small
\begin{tabular}{lrccc}
\toprule
age & $n$ & Correct\% & Halluc\% & Omit\% \\
\midrule
0   & 1438 & 59.9 & 33.0 & 7.1 \\
20  & 1052 & 39.0 & 49.8 & 11.2 \\
40  & 803  & 36.0 & 53.7 & 10.3 \\
60  & 593  & 33.6 & 57.2 & 9.3 \\
80  & 364  & 32.4 & 54.7 & 12.9 \\
100 & 178  & 32.6 & 51.7 & 15.7 \\
\bottomrule
\end{tabular}
\end{table}

Table~\ref{tab:app-cat-allage} gives per-category accuracy for all paradigms on the pooled all-age
set, the counterpart to the age-$0$ Table~\ref{tab:cat}. Read against the age-$0$ table it shows
where the ranking inverts: MemOS, flat with age, overtakes OPSD on the aged pooled set overall
($46.1$ vs $43.7$) and on most content categories, while OPSD keeps a clear lead only on Memory
Conflict ($57.2$ vs $47.7$). This is the honest scope condition on \S\ref{sec:main}: weight
memory's advantage is concentrated on recent material.

\begin{table}[t]
\centering
\caption{Per-category accuracy on the pooled all-age set (HaluMem-Long, $n{=}4428$), Correct\%.
Counterpart to the age-$0$ Table~\ref{tab:cat}.}
\label{tab:app-cat-allage}
\small
\setlength{\tabcolsep}{4pt}
\begin{tabular}{lccccccc}
\toprule
condition & Basic & Conflict & Dynamic & Multi-hop & General. & Boundary & \textbf{Overall} \\
\midrule
GT points \emph{(oracle)} & 76.5 & 73.2 & 60.6 & 43.9 & 33.5 & 75.6 & \textbf{63.4} \\
\midrule
Extracted (dump)   & 42.1 & 50.1 & 26.7 & 20.8 & 21.0 & 83.4 & 46.7 \\
MemOS (retrieval)  & 32.4 & 47.7 & 24.5 & 16.9 & 18.6 & 95.9 & 46.1 \\
Raw dialogue       & 31.1 & 41.5 & 18.9 & 10.2 & 17.5 & 69.5 & 37.3 \\
\midrule
OPSD (weight)      & 26.6 & 57.2 & 13.4 & 12.2 & 21.6 & 83.8 & 43.7 \\
LoRA $r$64 (persistent adapter) & 14.6 & 56.8 & 7.1 & 10.6 & 11.8 & 88.4 & 39.5 \\
\bottomrule
\end{tabular}
\end{table}

\subsection{Verify-then-regenerate: full breakdown}
\label{sec:app-vtr}

Table~\ref{tab:regen} gives VtR's per-category regeneration behavior, referenced from
\S\ref{sec:route}; the main text reports its retention (Figure~\ref{fig:age}).
Table~\ref{tab:app-vtr-age} gives the by-age accuracies behind those, and
Table~\ref{tab:app-vtr-gate} the gate accounting. VtR improves at every age including
age $0$, with the largest gains at ages $20$--$80$ where OPSD has forgotten but retrieval still
holds the fact. The gate keeps $22\%$ of answers as SUPPORTED and regenerates the rest; of the
$3434$ regenerated answers it fixes $814$ and breaks $277$ (net $+537$, a $21\%$ damage rate on the
already-correct answers it touched).

\begin{table}[t]
\centering
\caption{VtR routing behaviour by category (native protocol, $n{=}1438$). ``regen.'' is the share
of the category the gate sent to regeneration; $\Delta$ is VtR's gain over OPSD
(accuracies in Table~\ref{tab:cat}). VtR improves five of six categories and regresses only on
memory conflict, diagnosed in \S\ref{sec:route}.}
\label{tab:regen}
\small
\begin{tabular}{lrr}
\toprule
category & regen.\ & $\Delta$ vs.\ OPSD \\
\midrule
Memory boundary               & 100\% & $+17.7$ \\
Multi-hop inference           &  67\% & $+12.7$ \\
Generalization \& application &  66\% & $+9.3$ \\
Dynamic update                &  60\% & $+5.3$ \\
Basic fact recall             &  42\% & $+4.8$ \\
Memory conflict               &  36\% & $-9.9$ \\
\midrule
\textbf{Overall}              &  62\% & $+6.2$ \\
\bottomrule
\end{tabular}
\end{table}

\begin{table}[t]
\centering
\caption{VtR accuracy by question age (HaluMem-Long, all ages), Correct\%. VtR improves over both
single systems at every age.}
\label{tab:app-vtr-age}
\small
\begin{tabular}{lrcccr}
\toprule
age & $n$ & OPSD & MemOS & \textbf{VtR} & $\Delta$ vs OPSD \\
\midrule
0   & 1438 & 59.9 & 47.8 & \textbf{66.1} & $+6.2$ \\
20  & 1052 & 39.0 & 45.6 & \textbf{56.2} & $+17.2$ \\
40  & 803  & 36.0 & 42.2 & \textbf{51.6} & $+15.6$ \\
60  & 593  & 33.6 & 42.3 & \textbf{50.9} & $+17.4$ \\
80  & 364  & 32.4 & 40.1 & \textbf{48.4} & $+15.9$ \\
100 & 178  & 32.6 & 36.5 & \textbf{43.3} & $+10.7$ \\
\bottomrule
\end{tabular}
\end{table}

\begin{table}[t]
\centering
\caption{VtR gate accounting (pooled $n{=}4428$). The merged store folds superseded values, which
collapses false CONFLICT verdicts from $36\%$ (raw store) to $1\%$.}
\label{tab:app-vtr-gate}
\small
\begin{tabular}{lrr}
\toprule
gate outcome & count & OPSD$\rightarrow$final Correct\% \\
\midrule
SUPPORTED (kept)            & 994 (22\%)  & 61.8 $\rightarrow$ 65.6 \\
UNSUPPORTED $+$ CONFLICT (regen) & 3434 (77\%) & 38.5 $\rightarrow$ 54.1 \\
\midrule
regenerated, FIXED         & 814 & \multicolumn{1}{r}{} \\
regenerated, BROKE         & 277 & net $+537$, damage $21\%$ \\
\bottomrule
\end{tabular}
\end{table}

\subsection{Routed hybrids and oracle ceilings}
\label{sec:app-route}

Because the two memory systems fail on different questions, a per-question oracle that always chooses
the better arm reaches $69.3\%$ against OPSD's $59.9\%$ at age $0$; this is the prize the negative
routing signals of \S\ref{sec:route} could not claim. Table~\ref{tab:app-route} shows the practical
routed hybrids we ran fall well short of it. Routing stable facts to a LoRA and volatile facts to a
text arm never beats the best single non-oracle text method (MemOS at $46.1\%$ pooled), because the
LoRA loses to retrieval even on the stable slice it is handed, and a ground-truth stability oracle
for the route does not rescue it. The LLM router is a high-recall, low-precision volatile detector
(agreement with the ground-truth route $71.6\%$, volatile recall $99.6\%$), so it over-routes to
the text arm. This is why the main text pursues verification (which observes the produced answer)
rather than query-side routing.

\begin{table}[t]
\centering
\caption{Routed hybrids (HaluMem-Long, pooled $n{=}4428$), Correct\%. \%LoRA is the share of
questions answered by the weight arm. No routed hybrid beats plain MemOS ($46.1\%$); the oracle
ceiling is far above all of them.}
\label{tab:app-route}
\small
\begin{tabular}{lccc}
\toprule
routed variant & route & Correct\% & \%LoRA \\
\midrule
full-LoRA $r64$ + extraction & LLM route & 46.2 & 18.8 \\
full-LoRA $r64$ + extraction & GT route  & 46.0 & 48.3 \\
stable-LoRA + extraction     & GT route  & 43.5 & 48.3 \\
stable-LoRA + MemOS          & GT route  & 41.2 & 48.3 \\
\midrule
\emph{oracle (any arm correct, age $0$)} & --- & \emph{69.3} & --- \\
\bottomrule
\end{tabular}
\end{table}

\subsection{Retrieval recency is a binary cliff, not a gradient}
\label{sec:app-recency}

A natural hypothesis is that retrieval accuracy degrades gracefully with how stale the fetched fact
is. It does not. Once question age and category are controlled and Memory Boundary is excluded, the
graded session-distance correlation is negligible for all three paradigms
($|r|\le0.10$; the pooled correlation has an inverted sign purely from age and boundary confounds,
so we do not report it). The real effect is binary: whether \emph{any} retrieved fact comes from
the current session. Table~\ref{tab:app-recency} shows this separates the paradigms cleanly:
extracted memory collapses to $3\%$ without a current-session fact (near-definitional, since it
only sees the current session), OPSD gains $+21$ points, and MemOS is nearly flat at $+3$ points.
That flatness is exactly why MemOS overtakes weight memory at high age. The only graded signal that
survives is recency \emph{variance} (a wider spread of retrieved-fact ages predicts worse accuracy
by $-0.07$ to $-0.10$), a topic-drift/interference signature too weak to route on. These are
working-note numbers (recency builder over the retrieval logs) and support the retrieval-is-flat
reading of \S\ref{sec:char}.

\begin{table}[t]
\centering
\caption{Accuracy conditioned on whether any retrieved fact is from the current session
(within-age, non-boundary, $n{=}3378$), Correct fraction.}
\label{tab:app-recency}
\small
\begin{tabular}{lccc}
\toprule
method & current-session fact & no current-session fact & $\Delta$ \\
\midrule
Extracted (dump) & 0.419 & 0.032 & $+0.387$ \\
OPSD (weight)    & 0.452 & 0.245 & $+0.207$ \\
MemOS (retrieval)& 0.331 & 0.300 & $+0.031$ \\
\bottomrule
\end{tabular}
\end{table}

\subsection{Confidence signals for routing}
\label{sec:app-conf}

\S\ref{sec:route} reports that confidence in the generated answer is not a usable global routing
signal. This subsection gives the detail. Self-consistency (agreement across $K{=}5$ samples) is a
weak, category-dependent correctness signal that \emph{inverts} on Memory Boundary: boundary
accuracy rises with consistency up to the $95$--$99\%$ bin ($0.971$) then collapses to $0.404$ at
perfect consistency, because a deterministic wrong value (a middle name grabbed from the surname)
repeats perfectly while genuine abstentions vary in phrasing. Per-token confidence is orthogonal to
self-consistency ($r{=}-0.02$) and slightly stronger, but on boundary it is largely an answer-length
proxy ($\mathrm{corr}(\text{length}, \text{ptc\_mean}){=}0.65$): a scoring penalty on the first
token is diluted across a long abstention but dominates a one-word value. A single global
correctness probe therefore averages two opposite confidence-to-correctness mappings and lands near
chance (AUC $0.63$). The signal only works once the boundary regime is separated.

The clean detector is a length-constant one. Teacher-forcing a fixed six-token abstain string
(constant for every question, so length-confound-free by construction) and reading its probability
gives an answerability detector with AUC $0.944$ on the extraction baseline and, critically, AUC
$0.918$ when scored under the OPSD weight adapter, confirming the gate reads the model's
\emph{parametric} knowledge state, not merely context visibility. But this detector predicts
answerability, not correctness: against the event that the model is wrong it is at chance
($0.516$), and \texttt{margin\_lp}/\texttt{minp} confidence features are actively anti-predictive.
Routing on it yields a single operating point that gains only $+1.0$ point over OPSD at age $0$ and
turns net-negative pooled over ages, because once OPSD forgets the threshold degenerates to
"always route to text." Table~\ref{tab:app-conf} collects the AUCs. This is the evidence base for
the four negative signals of \S\ref{sec:route} and for why VtR verifies the produced answer instead.

\begin{table}[t]
\centering
\caption{Confidence-signal AUCs. The fixed-candidate abstain probe detects answerability well and
transfers to weight memory, but no signal predicts correctness. Extraction-baseline numbers from
the confidence-calibration note; OPSD transfer from the fixed-candidate note}
\label{tab:app-conf}
\small
\begin{tabular}{lccc}
\toprule
signal / target & all & non-boundary & boundary \\
\midrule
self-consistency $\rightarrow$ correct   & 0.606 & 0.631 & 0.391 (inverts) \\
per-token conf. $\rightarrow$ correct     & 0.628 & 0.855 & 0.931 \\
per-token conf., age-$0$ only            & 0.672 & 0.738 & 0.870 \\
$p_{\text{abs}}$ (fixed str.) $\rightarrow$ answerable, extraction & 0.944 & --- & --- \\
$p_{\text{abs}}$ under OPSD adapter $\rightarrow$ answerable       & 0.918 & --- & --- \\
$p_{\text{abs}}$ $\rightarrow$ OPSD \emph{wrong} (correctness)     & 0.516 & --- & --- \\
\bottomrule
\end{tabular}
\end{table}

\subsection{Weak categories are partly measuring the rubric}
\label{sec:app-weak}

Dynamic Update and Multi-hop Inference are our two smallest and lowest-scoring categories, and much
of that score reflects the evaluation rubric rather than the memory systems. On the $173$ age-$0$
questions in these two categories, $97$ (56\%) are failed by \emph{all five} methods including
full context, which sees the entire dialogue, so retrieval cannot be the dominant cause.
Table~\ref{tab:app-weak} buckets those $97$ universal failures: only about half are genuine content
errors, while $34\%$ have a gold answer not fully recoverable from the evidence the dataset itself
cites (Multi-hop golds are abstractive one-line summaries; only $48\%$ are fully reconstructible
from the cited evidence) and $11\%$ are cases where a method stated the gold verbatim and was still
scored wrong. Verbosity is punished: hallucination-judged answers carry a median of $28$ content
words against a $4$-word median gold, and $14.4\%$ of all hallucination verdicts contain at least
$80\%$ of the gold content plus extra true-but-unasked detail. Even the per-question oracle reaches
only $0.48$/$0.39$ here versus $0.74$--$0.98$ elsewhere. The honest reading is that roughly $45\%$
of all-method failures in these categories are rubric or gold artifacts, so the $0.24$--$0.30$
numbers should be read against an achievable ceiling of $0.39$--$0.48$, not as a system failure.
This rubric artifact is the reason we do not tune the router on these categories.

\begin{table}[t]
\centering
\caption{The $97$ age-$0$ Dynamic Update $+$ Multi-hop questions failed by all five methods,
bucketed by cause. Nearly half are gold or rubric artifacts rather than system defects.}
\label{tab:app-weak}
\small
\begin{tabular}{lrr}
\toprule
bucket & $n$ & share \\
\midrule
Genuine content error (all methods wrong differently) & 51 & 53\% \\
Under-specified gold (not recoverable from cited evidence) & 33 & 34\% \\
Judge strictness (gold stated verbatim, still marked wrong) & 11 & 11\% \\
Universal abstention (date-scoping) & 2 & 2\% \\
\bottomrule
\end{tabular}
\end{table}

% \subsection{A caveat that cuts against us: fine-mode retrieval is a wash}
% \label{sec:app-fine}

% We tested whether a richer MemOS-parity retriever (query goal-parsing plus multi-vector dense and
% sparse key/tag recall, our "fine" mode) improves over the single-vector "fast" retriever that our
% MemOS numbers use. It does not. On HaluMem-Medium ($4$ users $\times$ $20$ sessions, identical
% replayed extraction), fine scored $61.4\%$ QA against fast's $64.7\%$, a $3.3$-point drop that is
% \emph{not} significant (McNemar exact $p{=}0.24$; $12$ fast-only-correct vs $6$ fine-only-correct of
% $18$ discordant pairs), so the honest verdict is a wash trending negative. Hallucination moved the
% wrong way ($18.5\%\rightarrow20.7\%$), and there was one real regression, on Multi-hop Inference
% ($0.429\rightarrow0.238$), the category multi-vector recall was supposed to help most. The
% mechanism was candidate dilution at fixed top-$k$: fine surfaced about $48\%$ more candidates but
% still returned only $20$, so loosely-related expansion matches displaced tightly-relevant hits, at
% about $25\times$ the search cost. We report this because it bounds the strength of our MemOS
% baseline: a more elaborate retriever did not lift it, so the OPSD margin in \S\ref{sec:main} is not
% an artifact of an under-powered text system.

\subsection{Representation dominates age on HaluMem-Medium}
\label{sec:app-repr}

On HaluMem-Medium, where each user's full history ($\sim$160k tokens) still fits the context
window, the representation of memory matters far more than question age. Answering the same
cumulative questions from ground-truth points, extracted memory, and raw dialogue gives
$63.7\% > 57.6\% > 54.2\%$ overall, with accuracy nearly flat across age in every condition (no
long-context decay at this scale). The extraction-over-dialogue win is entirely a Memory Boundary
effect (dialogue $46.6\%$ vs extraction $72.6\%$): a clean extracted list lets the reader say
"unknown," whereas a raw transcript invites a hallucinated answer, and on every other category
extraction slightly loses to raw dialogue because extraction drops or fragments detail. This
localizes where compression helps (abstention) and where it hurts (fine detail), and it is why the
main results use the longer HaluMem-Long, where the context wall actually bites. from a cumulative-representation probe ($8$ users, $8630$ questions/condition)

\subsection{RouterMem-Mini: volatile facts are the hard axis}
\label{sec:app-rmm}

To probe volatile-versus-stable memory more directly than HaluMem allows, we built RouterMem-Mini,
a small generator ($8$ users $\times$ $20$ sessions) with an enriched volatile-fact axis
($173$ of $520$ questions typed volatile through explicit update chains) and genuinely unanswerable
Memory Boundary questions. The round-2 evaluation confirms the split that motivates routing: text
methods drop from $66$--$69\%$ on stable facts to $31\%$ on volatile facts, and the weight memory
drops from $29$--$31\%$ to $8\%$. Improved third-person minimal-span augmentation did not lift the
weight memory ($22$--$23\%$ overall, $r{=}100 \approx r{=}200$), which holds only on Memory Boundary
abstention ($92$--$94\%$), confirming that the roughly $20\%$ weight-memory ceiling on this data is a
weight-memory limit and not an augmentation-quality artifact. This corroborates the volatile-facts
finding of \S\ref{sec:char} on an independently generated benchmark.

\subsection{Where each memory system wins, qualitatively}
\label{sec:app-winpattern}

The one robust place text memory beats weight memory is absence. Analyzing the disjoint wins at
age $0$, the only category where either MemOS or extracted memory beats OPSD net is Memory Boundary:
of $120$ MemOS-only wins, $47$ are boundary questions where OPSD confabulates a value (its boundary
omission rate is $0.000$, so it never abstains, it invents), and the clearest sub-pattern is the
"precise unrecorded attribute" question (a salary, an exact weight, a film title the dataset never
recorded) where a text system reads its list, finds no such field, and reports the absence, while
OPSD produces a concrete number, inconsistently across ages for the same question. The remaining
non-boundary "wins" are mostly not a text-memory capability: about $18\%$ are OPSD over-answering
artifacts (it contained the gold but added extra true detail and was scored wrong) and a majority
are lenient passes where the text answer contained under half the gold content. We include this
because it is the honest mechanism behind the one category OPSD loses in \S\ref{sec:main} and
\S\ref{sec:absence}: inspectability, not knowledge. We also flag a judge inconsistency found here,
where roughly $20\%$ of OPSD's "hallucination" verdicts on boundary questions were correct
abstentions scored wrong while a near-identical text abstention was scored correct, so the stated
OPSD fabrication rate is inflated by about a fifth.

\subsection{LoCoMo second benchmark: full per-category breakdown}
\label{sec:app-locomo}
Table~\ref{tab:app-locomo} gives the per-category accuracy behind the overall comparison in
Figure~\ref{fig:locomo}. We port LoCoMo and use
its standard protocol (all questions asked once after the full conversation) over LoCoMo's five
categories, mapping its adversarial category to abstention (the answer is not in the conversation).
Weight memory is the pooled variant (all sessions distilled into a single adapter, since there is no
age axis). OPSD tracks extracted text memory across every content category and edges it overall
($46.9$ vs $46.4$); both, like all paradigms, fail on adversarial (unanswerable) questions, which
verify-then-regenerate repairs ($73.1$) to take the overall best ($60.4$). This mirrors the HaluMem
result---weight memory competitive with text on content, the router winning through recovered
abstention---on an independently-constructed benchmark.

\begin{table}[t]
\centering
\small
\caption{LoCoMo end-of-conversation accuracy (Correct\%) by question category (Qwen3.5-9B,
Claude-Sonnet judge, $1{,}586$ questions). Best per column in bold.}
\label{tab:app-locomo}
\begin{tabular}{lcccccc}
\toprule
Method & Single-hop & Temporal & Multi-hop & Open-dom. & Adversarial & Overall \\
\midrule
Raw dialogue (full-context) & \textbf{74.7} & 7.2 & 30.1 & 31.2 & 2.5 & 39.1 \\
Extracted (text) & 69.3 & \textbf{61.7} & 33.3 & \textbf{32.3} & 3.4 & 46.4 \\
MemOS (retrieval) & 46.1 & 40.2 & 18.8 & 31.2 & 4.7 & 31.3 \\
\midrule
SFT (weight) & 23.7 & 10.3 & 13.1 & 14.6 & 11.7 & 16.9 \\
OPSD (weight) & 70.2 & 61.3 & \textbf{33.5} & 25.0 & 10.6 & 46.9 \\
\midrule
VtR (weight\,+\,text) & 69.2 & 57.4 & 28.6 & 21.1 & \textbf{73.1} & \textbf{60.4} \\
\bottomrule
\end{tabular}
\end{table}

%%%%%%%%%%%%%%%%%%%%%%%%%%%%%%%%%%%%%%%%%%%%%%%%%%%%%%%%%%%%

%%% checklist for neurips requirement. you can comment it out if not needed.
%\newpage
%\input{checklist.tex}

\end{document}